\documentclass[letterpaper, 10pt, conference]{ieeeconf}      

\IEEEoverridecommandlockouts                              

\usepackage[margin=0.75in]{geometry}

\usepackage{amsmath} 
\usepackage{amssymb}  
\usepackage{bm}

\usepackage{graphicx}
\usepackage[colorlinks=true, allcolors=blue]{hyperref}
\usepackage[official]{eurosym}

\usepackage{textcomp}
\usepackage{caption}
\usepackage{booktabs}

\usepackage{multirow}
\usepackage{subcaption}
\usepackage[percent]{overpic}
\usepackage{hyperref}

\newcommand{\HandName}{ISyHand}
\title{\vspace{1cm}\LARGE \bf
\HandName: A Dexterous Multi-finger Robot Hand\\ with an Articulated Palm}

\author{Benjamin A. Richardson$^{*,1}$, Felix Grüninger$^{*,2}$, Lukas Mack$^{*,3}$, Joerg Stueckler$^{3}$, Katherine J. Kuchenbecker$^{1}$
\thanks{$^*$ These authors contributed equally.}%
\thanks{$^{1}$ Benjamin A. Richardson and Katherine J. Kuchenbecker are with the Haptic Intelligence Department, Max Planck Institute for Intelligent Systems, Stuttgart, Germany,
       {\tt\small \{richardson, kjk\}@is.mpg.de}}%
\thanks{$^{2}$ Felix Gr{\"u}ninger is with the Robotics ZWE, Max Planck Institute for Intelligent Systems, Stuttgart and Tübingen, Germany,
       {\tt\small grueninger@is.mpg.de}}%
\thanks{$^{3}$ Lukas Mack and Joerg Stueckler are with the Intelligent Perception in Technical Systems Group, University of Augsburg, Augsburg, Germany,
       {\tt\small \{lukas.mack, joerg.stueckler\}@uni-a.de}}%
}

\usepackage{tikz}
\newcommand\acceptedtext{%
  \footnotesize Accepted for publication at the IEEE-RAS International Conference on Humanoid Robots (Humanoids) 2025.}

\newcommand\acceptednotice{%
\begin{tikzpicture}[remember picture,overlay]
\node[anchor=north,yshift=-10pt] at (current page.north) {\parbox{\dimexpr0.75\textwidth-\fboxsep-\fboxrule\relax}{\acceptedtext}};
\end{tikzpicture}%
}

\newcommand\copyrighttext{%
  \footnotesize \textcopyright 2025 IEEE. Personal use of this material is permitted.
  Permission from IEEE must be obtained for all other uses, in any current or future
  media, including reprinting/republishing this material for advertising or promotional
  purposes, creating new collective works, for resale or redistribution to servers or
  lists, or reuse of any copyrighted component of this work in other works.
  }
\newcommand\copyrightnotice{%
\begin{tikzpicture}[remember picture,overlay]
\node[anchor=south,yshift=10pt] at (current page.south) {\fbox{\parbox{\dimexpr\textwidth-\fboxsep-\fboxrule\relax}{\copyrighttext}}};
\end{tikzpicture}%
}

\begin{document}

\maketitle
\acceptednotice
\copyrightnotice
\thispagestyle{empty}
\pagestyle{empty}

\begin{abstract}
The rapid increase in the development of humanoid robots and customized manufacturing solutions has brought dexterous manipulation to the forefront of modern robotics. 
Over the past decade, several expensive dexterous hands have come to market, but advances in hardware design, particularly in servo motors and 3D printing, have recently facilitated an explosion of cheaper open-source hands.
Most hands are anthropomorphic to allow use of standard human tools, and attempts to increase dexterity often sacrifice anthropomorphism. 
We introduce the open-source \HandName\ (pronounced easy-hand), a highly dexterous, low-cost, easy-to-manufacture, on-joint servo-driven robot hand. Our hand uses off-the-shelf Dynamixel motors, fasteners, and 3D-printed parts, can be assembled within four hours, and has a total material cost of about 1,300\,USD. The \HandName's unique articulated-palm design increases overall dexterity with only a modest sacrifice in  anthropomorphism. To demonstrate the utility of the articulated palm, we use reinforcement learning in simulation to train the hand to perform a classical in-hand manipulation task: cube reorientation. 
Our novel, systematic experiments show that the simulated \HandName\ outperforms the two most comparable hands in early training phases, that all three perform similarly well after policy convergence, and that the \HandName\ significantly outperforms a fixed-palm version of its own design. 
Additionally, we deploy a policy trained on cube reorientation on the real hand, demonstrating its ability to perform real-world dexterous manipulation. 



\end{abstract}

\section{Introduction}
The dexterity, strength, robustness, and tactile sensing of the human hand are crucial to the human ability to perceive, manipulate, and use objects. A person can firmly squeeze and operate a heavy tool in one hand while the other gently holds a delicate object. This vast range of capabilities is crucial to the human ability to physically interact with objects and accomplish tasks across a wide array of environments. Although reproducing the aforementioned traits in a robot hand is challenging, human-like dexterous manipulation will be necessary if robots are to become more ubiquitous and generally capable in human-centered environments. 

\begin{figure}[h]
    \centering
    \includegraphics[width=0.90\columnwidth, trim={0.1cm 1.3cm 0.1cm 1.5cm}, clip]{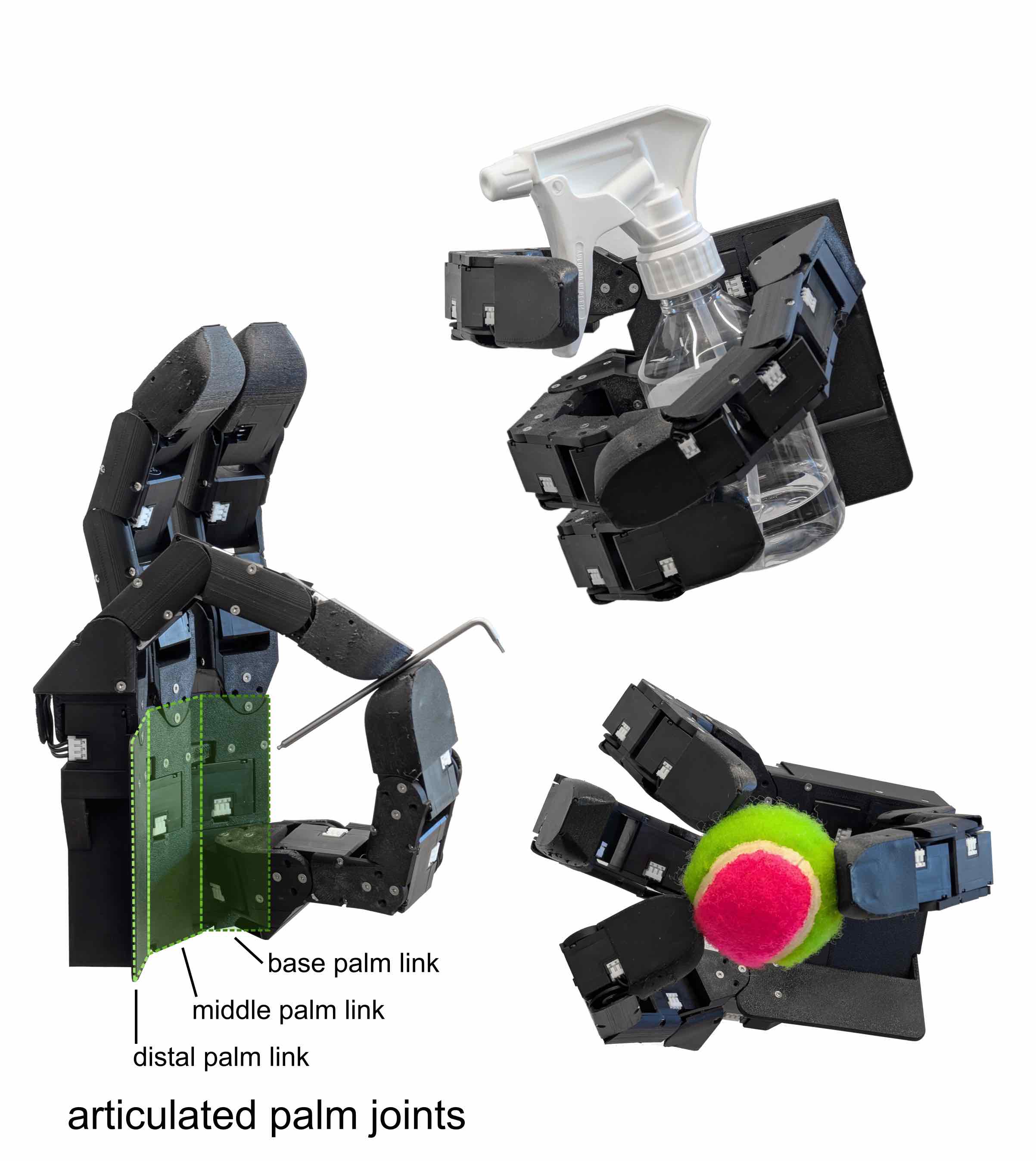}
    \caption{The \HandName's articulated palm facilitates grasping.}
    \label{fig:ourhand}
    \vspace{-13pt}
\end{figure}

\subsubsection*{Robotic Grippers}
Some work has pushed simple grippers, such as parallel jaw or underactuated multi-finger grippers, to their manipulation limits \cite{ma2017yale,right_hand,spiers2018variable,shehawy2023flattening,xie2023hand}, but these grippers are fundamentally limited in their dexterity. Parallel jaw grippers can perform pick-and-place tasks, pinch laundry and fold it on a hard surface \cite{shehawy2023flattening}, and even perform simple object rotations around a single axis \cite{spiers2018variable, xie2023hand}. Underactuated multi-fingered grippers typically excel at grasping by conforming to the shape of a grasped object, but they lack the dexterity for robust manipulation \cite{ma2017yale,right_hand}. The most dexterous grippers are multi-finger hands with fully actuated or almost fully actuated degrees of freedom (DoF). 

High-DoF multi-finger hands are more dexterous, but this capability comes at the cost of needing many individual actuators to move each joint independently and coordinate finger movements for more complex manipulation tasks. Several different actuation mechanisms have been implemented on multi-fingered hands; these are primarily tendon-driven \cite{shadow_hand,christoph2025orca} and on-joint servo-driven \cite{allegro,HDHM2023,shaw2023leaphand,xu2023dexterous,H4ND2024}, although pneumatically actuated hands have also been developed \cite{festo,deimel2016novel}. 
Tendon-driven hands provide certain advantages: they are similar in size to a human hand, can move very quickly, and can generate high torque at certain joints by using multiple tendons. However, they are typically quite complex to manufacture and maintain, and most of the actuators need to be located outside the hand, typically inside a large forearm. They can also be more difficult to control because of the interactions between tendons, the need to calibrate motor positions, and their complex forward dynamics. 
Pneumatically driven hands have similar drawbacks, with large wrists for valves, intricate tubing to direct airflow, and likely an external cabinet to supply pressurized air. They can also be difficult to control, requiring precise valves to accurately control pressure, and their actuation can be noisy. 
On-joint servo-driven hands offer a balanced approach to design and control. While typically larger than tendon-based and pneumatic hands, these hands are much simpler to manufacture, maintain, and repair. Additionally, they are far easier to control; each joint can simply be commanded to a desired position. Servo-driven hands therefore seem to be the most accessible option for the research community.

\begin{figure}[tb!]
    \centering
    \vspace{1ex}
    \includegraphics[width=0.85\columnwidth, trim={0cm 1.75cm 0cm 1.75cm}, clip]{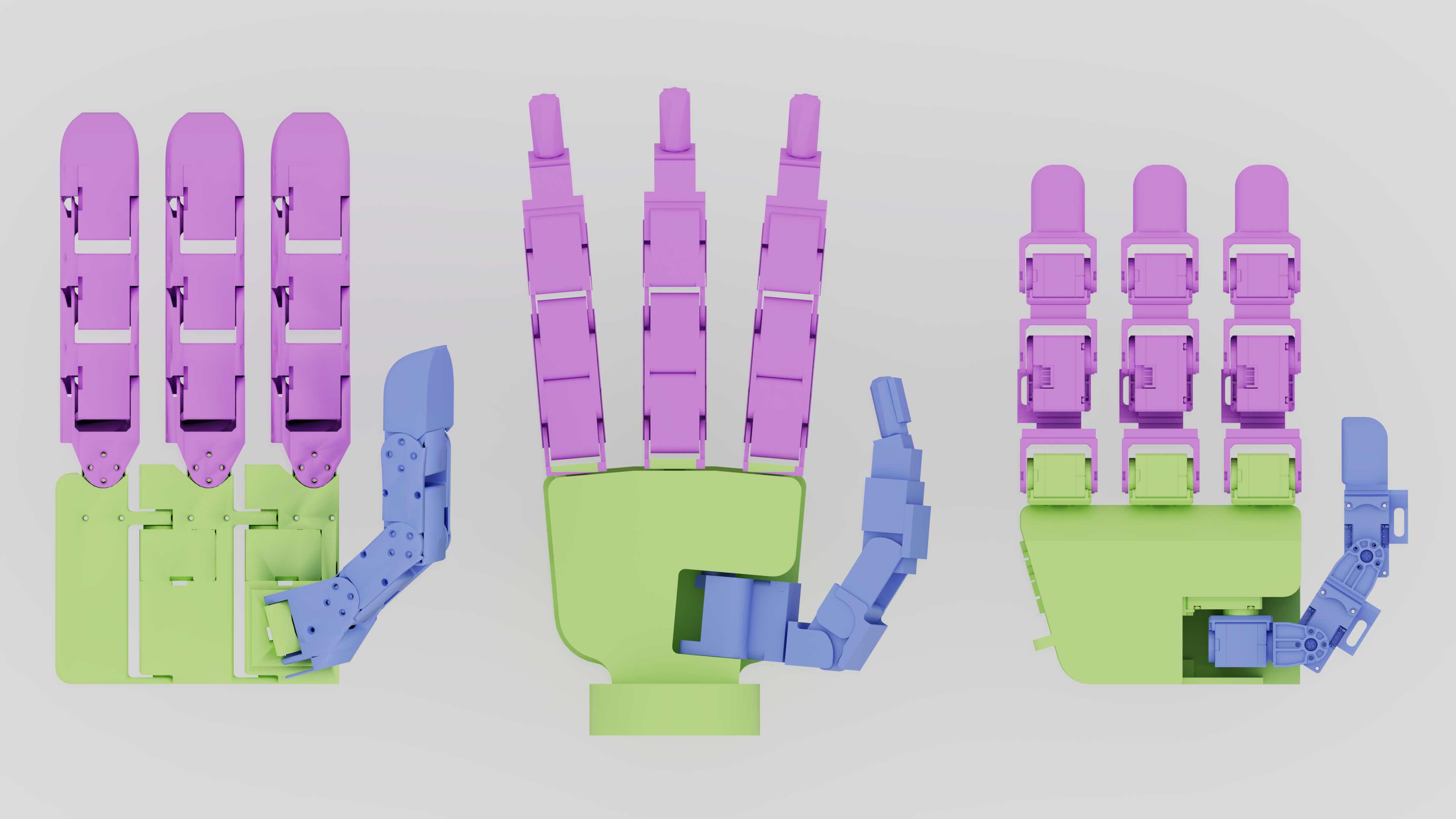}
    \caption{Size comparison between our \HandName\ (left), the Allegro hand (center), and the LEAP hand (right).}
    \label{fig:all_hands}
    \vspace{-10pt}
\end{figure}

Two of the most prominent servo-driven multi-finger robot hands are the Allegro hand and the LEAP hand~\cite{shaw2023leaphand}, both of which are similar in size to our \HandName\ (see Figs.~\ref{fig:ourhand}, \ref{fig:all_hands}).
The servo-driven, four-fingered, 16-DoF Allegro hand is a popular tool for robotic in-hand manipulation \cite{handa2023dextreme,wang2024penspin} likely because it is easy to use and not prohibitively expensive (17,000\,USD). 
However, the custom closed-source hardware makes it difficult to repair, and hardware failures have been repeatedly reported \cite{shaw2023leaphand,handa2023dextreme}, although the new version of the hand uses more robust motors.
The Tilburg hand is very similar to the Allegro Hand, though less expensive (5,170\,USD) \cite{tilburg}. However, it is still closed-source.
The LEAP hand has four fingers with 16 DoF and addresses some of the shortcomings of the Allegro hand by introducing a novel kinematic mechanism in the fingers that greatly increases finger dexterity when the hand is open or closed. Because it is fully open-sourced and can be assembled from off-the-shelf components, it can be built and repaired by a larger group of non-experts. Additionally, it is substantially less expensive, with a cost of about 2,000\,USD. However, the finger kinematics make teleoperation awkward because they are highly non-anthropomorphic. Additionally, it is unclear if the LEAP hand excels at other manipulation tasks. Finally, because it uses off-the-shelf connectors to link the motors, the design is not easy to modify or customize. 

We propose the \HandName\ (pronounced \textit{easy-hand}), a robust, dexterous, 18-DoF robot hand (Fig.~\ref{fig:ourhand}). Like the LEAP hand, the \HandName\ is inexpensive and made with off-the-shelf and 3D-printed components. It takes approximately four hours to assemble with a material cost of approximately 1,300\,USD. Because its body and linkages are 3D-printed, our hand can easily be repaired and broken parts replaced. The \HandName\ uses a unique linkage design that routes and houses cabling along the sides of the fingers, reducing wear and risk of damage. Additionally, we introduce a new kinematic mechanism, a 2-DoF articulated palm, that substantially increases the dexterity of the \HandName\ while allowing us to maintain anthropomorphism in the finger joints.\looseness-1



\subsubsection*{In-hand Manipulation}
Besides evaluating the low-level kinematic and dynamic properties of a robot hand and its actuators (joint repeatability/endurance, thumb opposability, finger manipulability ellipsoids, and grasp strength),
related work also demonstrates more complex manipulation tasks to highlight possible applications for particular robot hands.
For example, manipulation dexterity has often been demonstrated by manually teleoperating the robot hand to grasp and interact with a variety of everyday objects \cite{zorin2025-RUKA,christoph2025orca,shaw2023leaphand}.
When recorded, these demonstrations can be used to distill autonomous manipulation policies through imitation learning that can solve the demonstrated tasks.
Christoph et al.~\cite{christoph2025orca} use teleoperated robot hand trajectories of a pick-and-place task to train an imitating policy that is later evaluated in repeated task execution over seven hours to showcase the real-world durability and repeatability of their robot hand.
Shaw et al.~\cite{shaw2023leaphand} pretrain manipulation policies for the LEAP hand with imitation learning on internet video data and finetune them with trajectories originating from the Allegro hand; in particular, they train for different pick, place, rotate, and push tasks and evaluate the ratio of successful task executions over multiple trials for both hands. 

Since collecting demonstrations for imitation learning is tedious, reinforcement learning (RL) can instead be deployed to train a policy from scratch in simulation and later transfer it to the real robot.
Shaw et al.~\cite{shaw2023leaphand} train an RL policy to rotate a cube in-hand around the vertical z-axis for both the LEAP and the Allegro hands in simulation and transfer the LEAP hand's policy onto the real system.
Similarly, Christoph et al.~\cite{christoph2025orca} train a policy to rotate a tennis ball in-hand.
Recent work demonstrates impressive RL policies for in-hand reorientation of objects, such as a medium-sized cube \cite{handa2023dextreme}, cylinders \cite{wang2024penspin}, and various small items \cite{qi2023rotateit}, all transferred to the real Allegro hand. While the aforementioned approaches lead to impressive showcases of dexterous manipulation, they mainly reflect the capabilities of the specific robot hand in combination with the particular machine-learning method. 
There is often no systematic evaluation of how the proposed contributions (e.g., new mechanical design features) influence a hand's performance in complex object-manipulation tasks in comparison to existing robots hands.

We propose a systematic approach to evaluate the performance of our new robot hand \HandName\ against the Allegro and LEAP hands in the prevalent in-hand cube reorientation task.
Specifically, for each robot hand we train for a fixed number of steps a set of RL policies that covers a grid of possible initial and target cube positions (see Section \ref{subsec:cube_reorient}).
This strategy enables us to systematically discover the best spatial cube manipulation location for each hand (and thus report the actual best performance for each hand) in early training phases and discover how manipulation performance varies across each hand's workspace. We further train the best policies for much longer to compare training progress and convergence.
Importantly, we perform an ablation study of our robot hand without its articulated palm joints, called \HandName\ (flat). Our grid evaluation allows us to demonstrate that palm articulation confers a substantial performance advantage for the in-hand cube reorientation task in early training phases. The open-source hand design and code can be found at \href{https://isyhand.is.mpg.de/}{isyhand.is.mpg.de}.

\begin{figure}[b!]
    \centering

    \begin{subfigure}[t]{0.45\linewidth}
        \centering
        \begin{overpic}[width=\linewidth, trim={12cm 4cm 25cm 0cm}, clip]{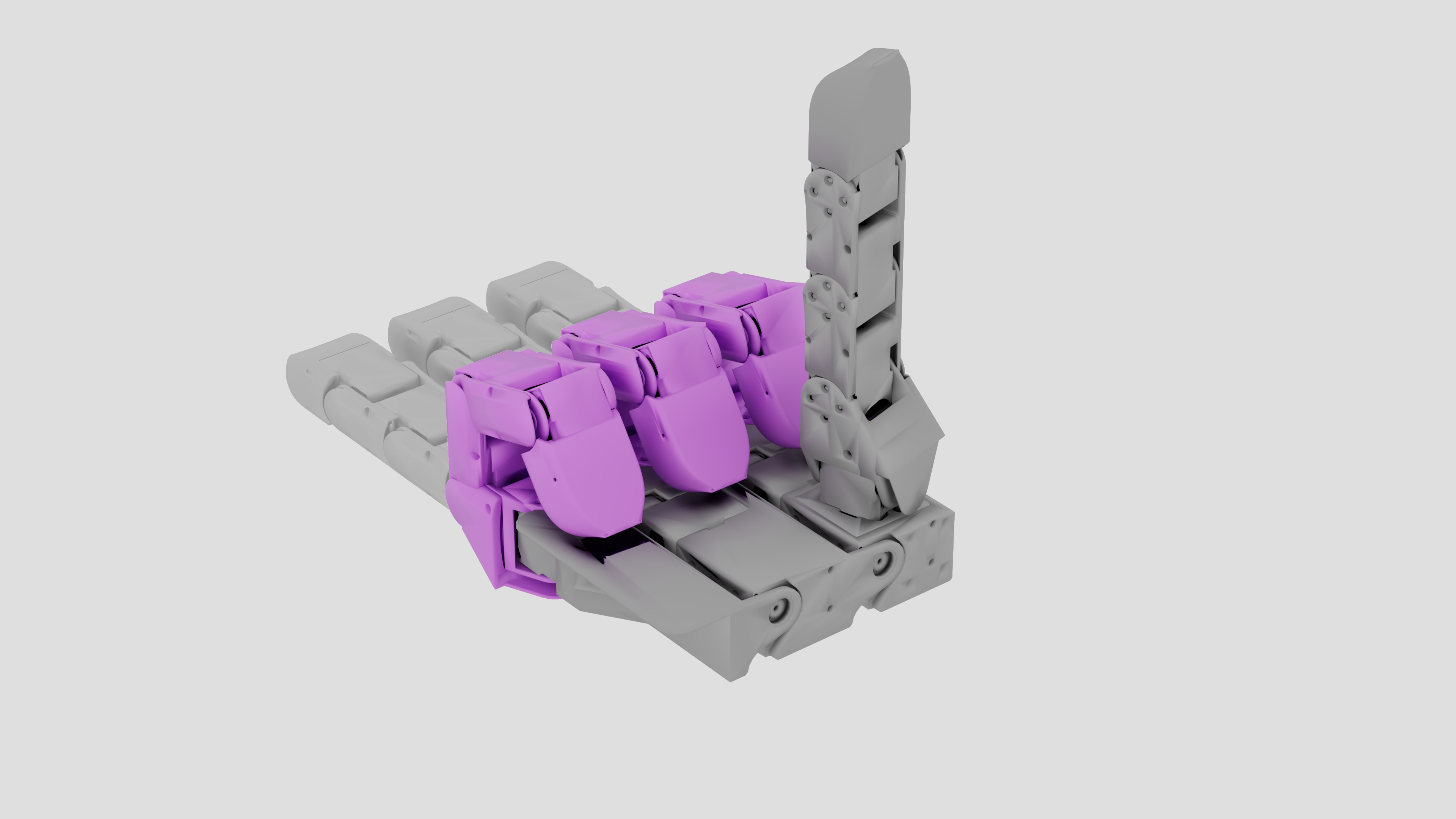}
            \put(5,62){\large\textbf{a}}
        \end{overpic}
    \end{subfigure}
    \begin{subfigure}[t]{0.45\linewidth}
        \centering
        \begin{overpic}[width=\linewidth, trim={12cm 4cm 25cm 0cm}, clip]{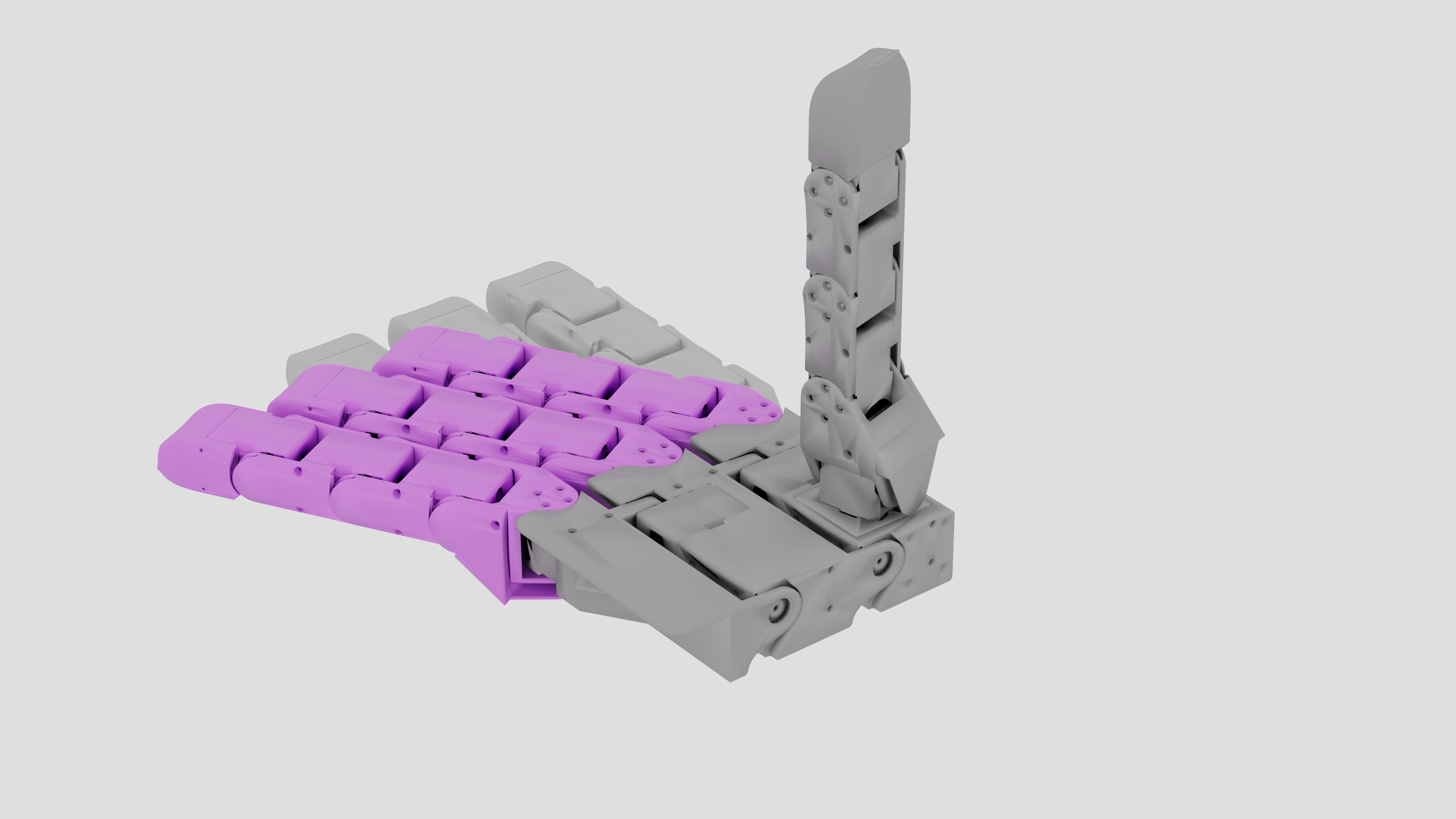}
            \put(5,62){\large\textbf{b}}
        \end{overpic}
    \end{subfigure}

    \vspace{0.3em}

    \begin{subfigure}[t]{0.45\linewidth}
        \centering
        \begin{overpic}[width=\linewidth, trim={12cm 4cm 25cm 0cm}, clip]{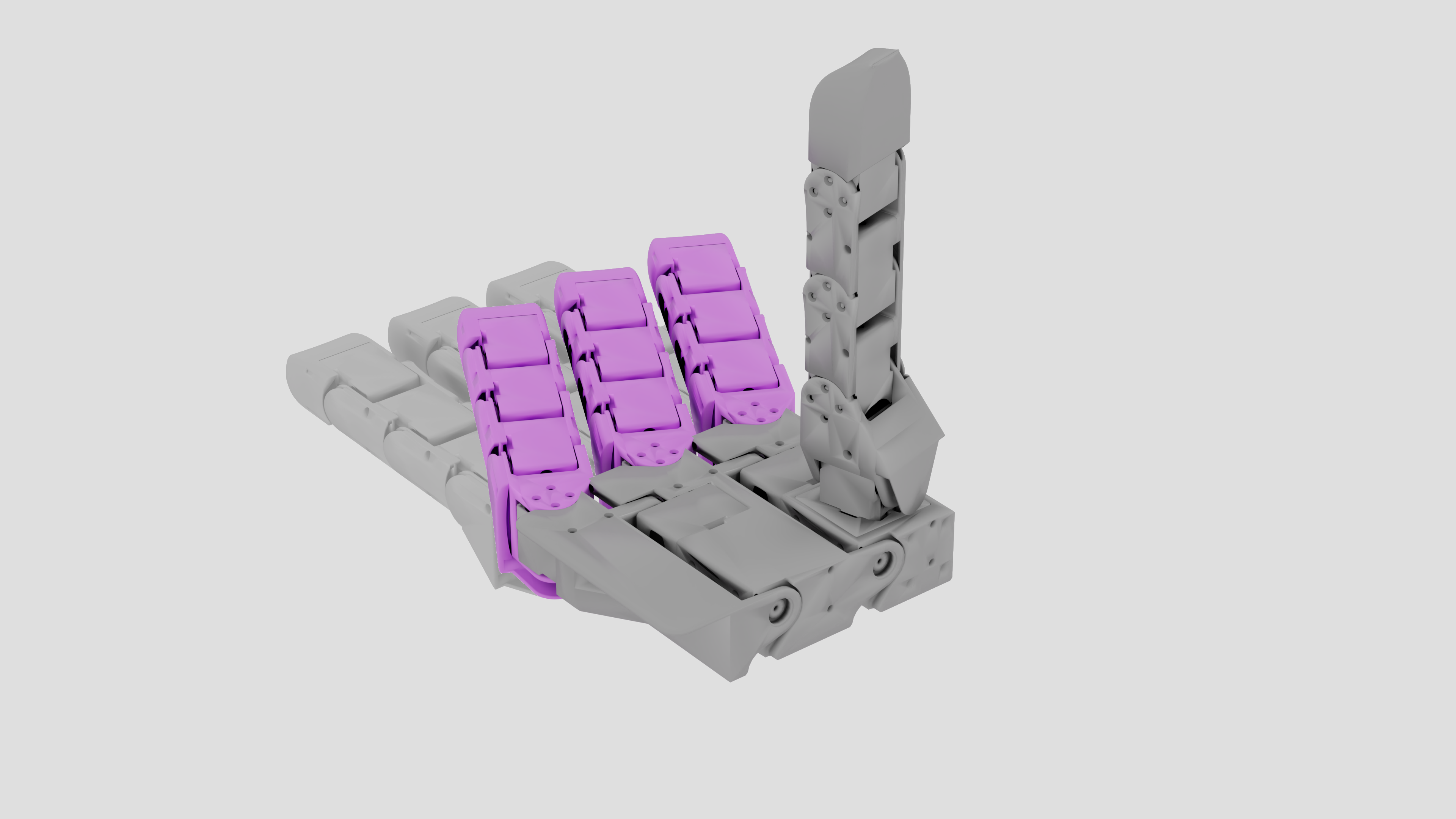}
            \put(5,62){\large\textbf{c}}
        \end{overpic}
    \end{subfigure}
    \begin{subfigure}[t]{0.45\linewidth}
        \centering
        \begin{overpic}[width=\linewidth, trim={12cm 4cm 25cm 0cm}, clip]{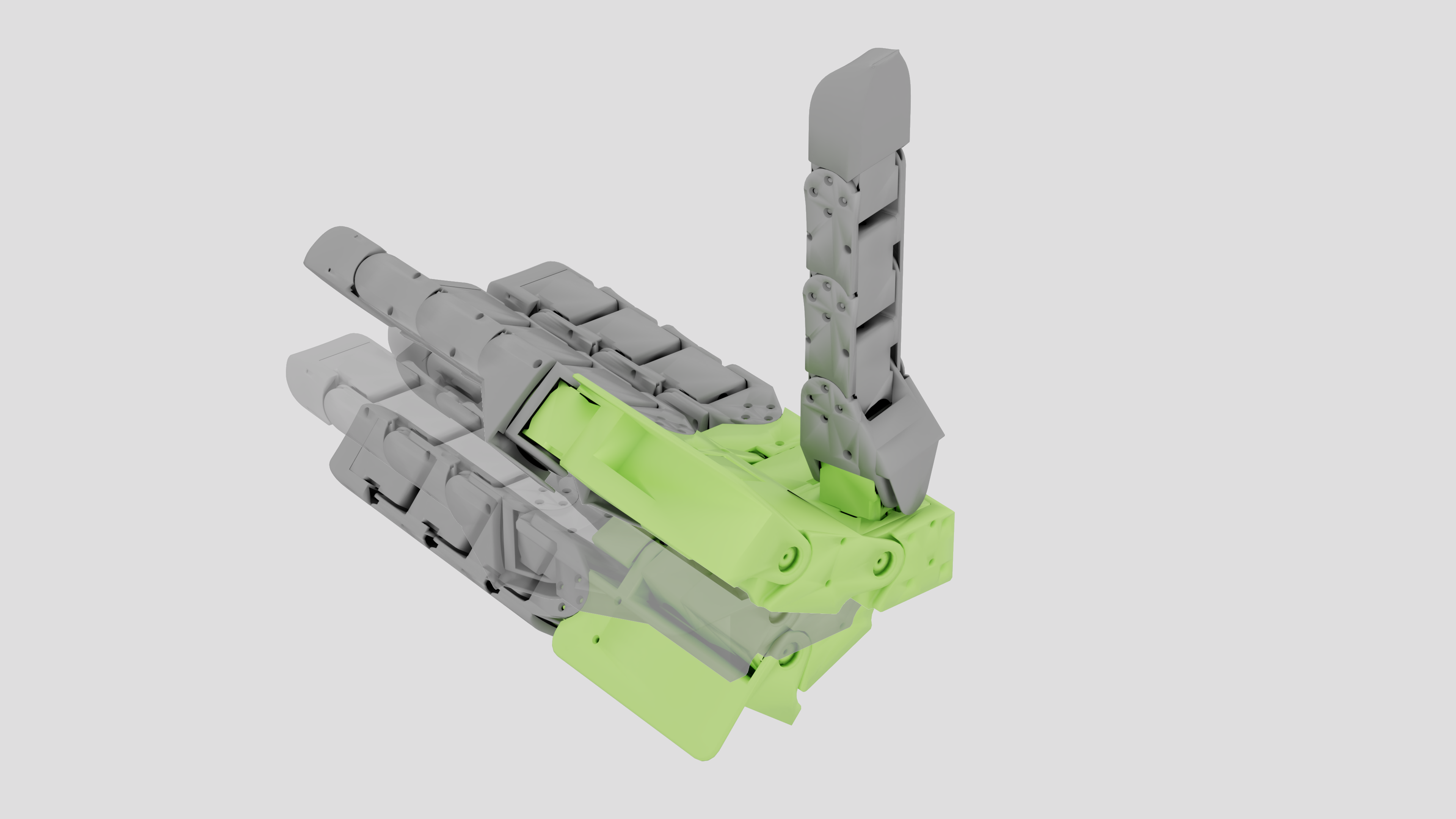}
            \put(5,62){\large\textbf{d}}
        \end{overpic}
    \end{subfigure}

    \vspace{0.3em}

    \begin{subfigure}[t]{0.45\linewidth}
        \centering
        \begin{overpic}[width=\linewidth, trim={12cm 4cm 25cm 0cm}, clip]{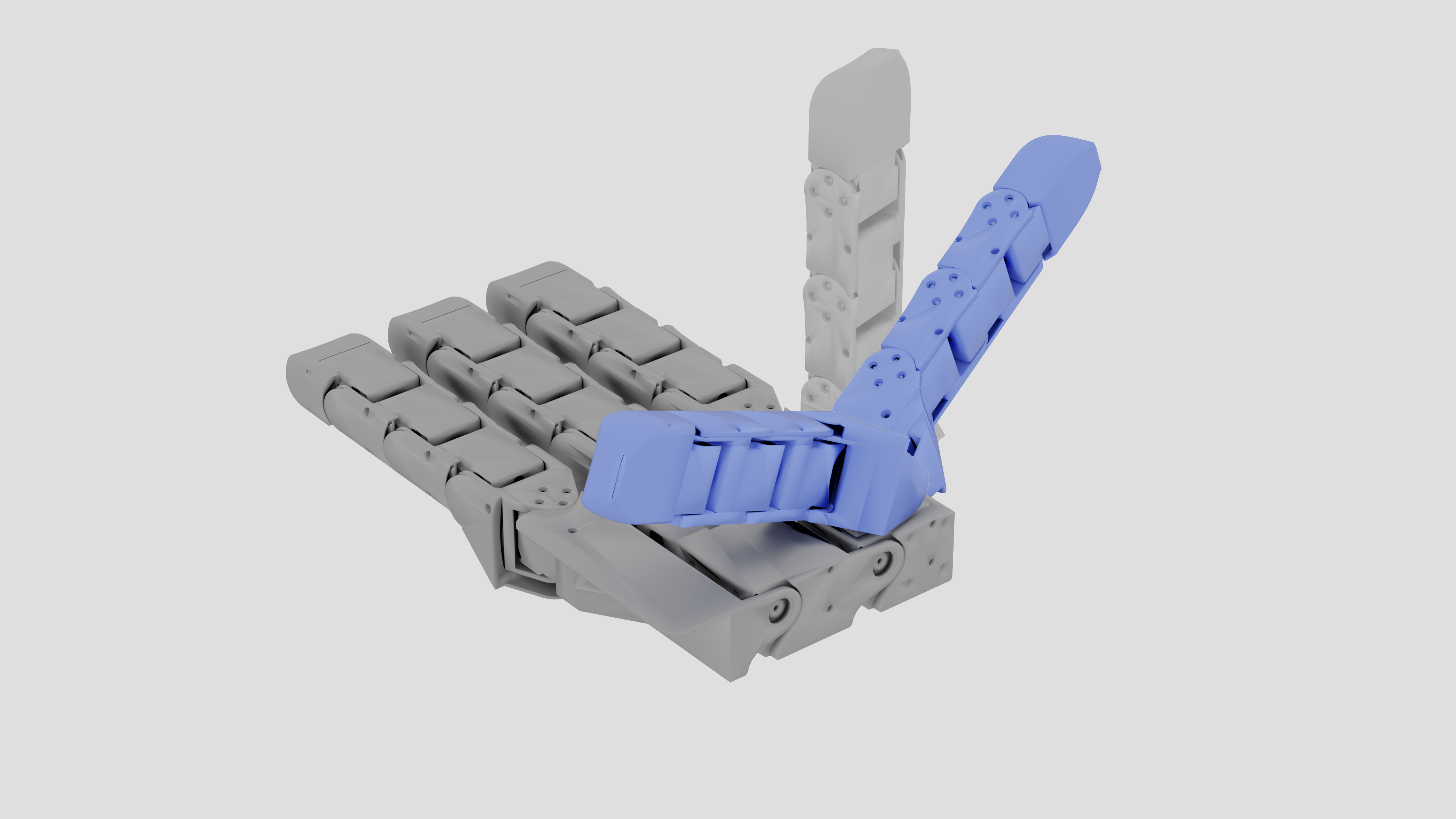}
            \put(5,62){\large\textbf{e}}
        \end{overpic}
    \end{subfigure}
    \begin{subfigure}[t]{0.45\linewidth}
        \centering
        \begin{overpic}[width=\linewidth, trim={12cm 4cm 25cm 0cm}, clip]{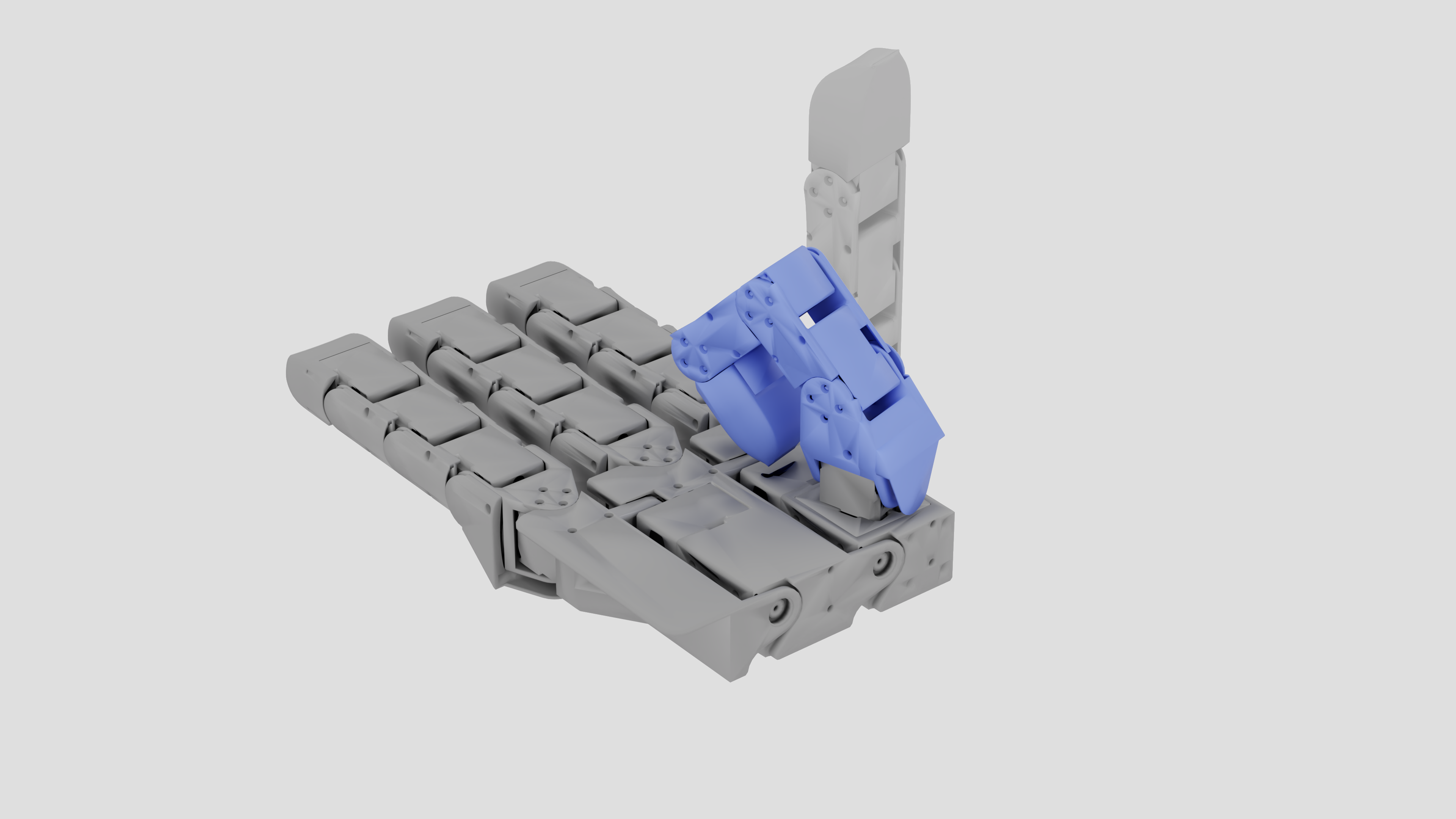}
            \put(5,62){\large\textbf{f}}
        \end{overpic}
    \end{subfigure}

    \caption{Six motions that display the 18 articulated DoF of the \HandName.
        (\textbf{a}) Three DoF in each finger allow for flexion.
        (\textbf{b, c}) Each finger is connected to the palm via a revolute joint that enables ab- and adduction.
        (\textbf{d}) Two articulated palm joints create a flexible palm.
        (\textbf{e}) The joint that connects the thumb to the palm enables thumb rotation.
        (\textbf{f}) Like the other fingers, the thumb has three DoF for flexion.
        While the finger flexions depicted in \textbf{a} and \textbf{f} are limited by self-contacts, the other articulations each show the full range of motion.} 
    \label{fig:dof}
\end{figure}

\section{Hand Design}
\begin{figure*}
    \centering
    \includegraphics[width=0.90\textwidth,trim={0.1cm 5.5cm 0.1cm 1cm},clip]{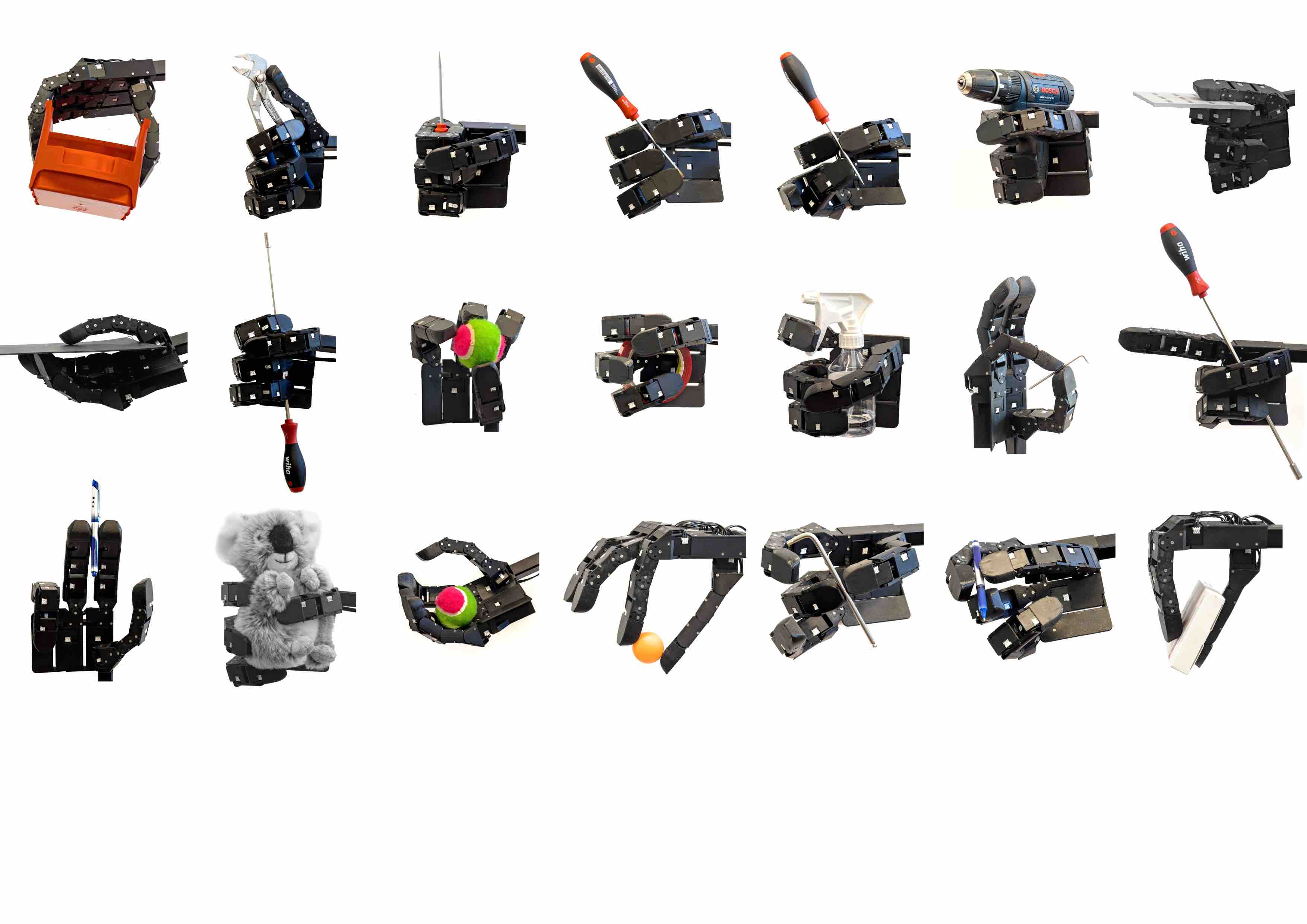}
    \vspace{-55pt}
    \caption{The \HandName\ performing a subset of grasps from the GRASP taxonomy~\cite{feix2016grasp}.}
    \label{fig:grasptaxonomy}
    \vspace{-12pt}
\end{figure*}

The \HandName\ resembles human hands and is approximately 50\% larger than the hand of an average male adult. At 255\,mm long, 130\,mm wide and 38\,mm thick, it is similarly sized to both the Allegro and LEAP hands and weighs approximately 620\,g.
%
%
%
Like many existing on-joint servo-driven hands, the \HandName\ has three fingers and a thumb. \textit{The main feature that sets our design apart from existing hands is its 2-DoF articulated palm.} This articulated palm allows the \HandName\ to mimic the degrees of freedom in human finger joints while increasing the system's overall dexterity compared to existing rigid-palm designs like the Allegro and LEAP hands. In total, the hand has 18 DoF, as illustrated in Fig.~\ref{fig:dof}. 
Each joint is driven by a Dynamixel motor, which provides high angular resolution and torque-limited position control. As demonstrated by the 21 diverse grasps shown in Fig.~\ref{fig:grasptaxonomy}, its articulated palm allows the \HandName\ to perform delicate pinching grips with the fingertips (e.g., plucking berries, grabbing very small objects), spread the fingers wide to grab larger objects, and cup smaller objects with a tighter grip.
Beyond the additional dexterity introduced by the palm joints, the \HandName\ was designed to be affordable, robust, modular, and a general platform for tactile sensing. The cost of all of the hand components is comparable to the LEAP hand and far cheaper than the Allegro or Tilburg hands. Because all components were custom-designed and are open-sourced, they can be modified to create additional morphologies. Additionally, the thumb attaches to the palm using an insert that can be customized to change the angle of the thumb for different applications. The palm and finger surfaces are flat for mounting thin tactile sensors, the fingertips can be easily modified to accommodate fingertip tactile sensors, and the cable routing was designed with enough space to accommodate additional wiring.

\subsection{Hardware}
The \HandName\ is assembled from Dynamixel motors, off-the-shelf fasteners, and 3D-printed linkages and soft fingertips. The total hardware cost (including the motors, motor electronics, power supply, fasteners, and 3D-printing filament) is approximately 1,300\,USD. A summary of the primary hardware components is shown in Table~\ref{tbl:hardware}. 

\begin{table}[b]
\vspace{-5pt}
\caption{Primary hardware components.}
\centering
\begin{tabular}{cccc}
\toprule
\textbf{Component} & \textbf{Model} & \textbf{Provider} & \textbf{Total Cost} \\ 
\midrule
\multirow{2}{*}{Servo Motors}    & 12x XL330-M288-T   & Robotis   & 330\,USD     \\  \vspace{3pt}
                                 & 6x XC330-M288-T    & Robotis   & 620\,USD    \\ \vspace{3pt}
Communication & U2D2 & Robotis & 60\,USD \\ \vspace{3pt}
Power & \begin{tabular}[c]{@{}c@{}}Custom injector,\\5V supply\end{tabular} & N/A           & 40\,USD                \\  \vspace{3pt}
Linkages   & CPE HG100      & Fillamentum   & 35\,USD \\ 
Fingertips & Filaflex 60A   & Recreus  & 60\,USD  \\ 
\bottomrule
\end{tabular}
\label{tbl:hardware}
\end{table}

\subsubsection{Motors}
The hand uses twelve Dynamixel XL330-M288-T and six XC330-M288-T motors. There are three XL motors in each finger and the thumb used for flexion and extension (Fig.~\ref{fig:dof}\textbf{a},\textbf{f}). The more powerful XC motors are used for finger ab- and adduction (Fig.~\ref{fig:dof}\textbf{b},\textbf{c}), palm flexion (Fig.~\ref{fig:dof}\textbf{d}), and thumb rotation (Fig.~\ref{fig:dof}\textbf{e}). The motors are daisy-chained along each link to a central hub on the back of the palm (Fig.~\ref{fig:isyhand_collage}). The off-the-shelf Robotis U2D2 is used for communication, and a custom board is used for power injection. The motors are controlled via USB using either the Dynamixel Wizard application or the Dynamixel SDK.

\subsubsection{3D-printed Components}
The majority of components other than the motors, specifically the palm segments and finger linkages, are 3D-printed plastic parts that can be fabricated on most 3D printers. The parts are designed to be robust when printed in a variety of materials, including Onyx, PCTG, PLA, PLA+, and flexible TPU. Additionally, the finger linkages are designed to symmetrically distribute forces and torques across the motor axes, reducing asymmetric loading and increasing the longevity of the motors. The fingertips are printed using Filaflex 60A to make them deformable for easier object manipulation, and they have embedded fingernails for precision pinching grasps. 

\subsection{Cabling}
\begin{figure}[t]
    \centering
    \includegraphics[width=0.45\textwidth]{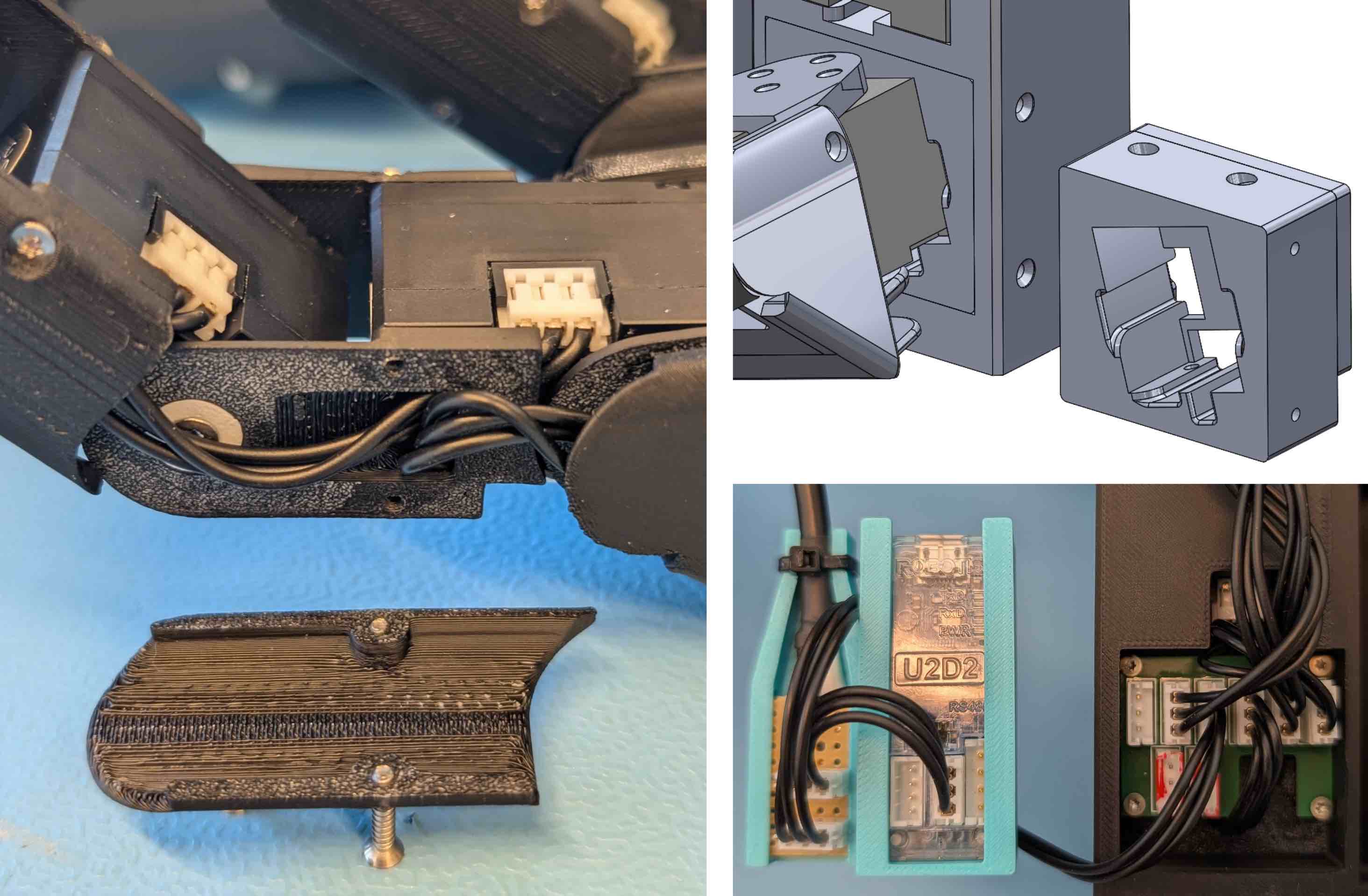}
    \caption{Left: The \HandName\ cable covers streamline routing of the three-wire motor cables, which are daisy-chained along each finger.
    Top right: Thumb adapter.
    Bottom right: Power injection and distribution.}
    \label{fig:isyhand_collage}
    \vspace{-12pt}
\end{figure}

One of the drawbacks of the LEAP hand and other existing on-joint servo-driven hands is the exposed motor wiring along the fingers, which often causes practical problems. A snagged wire can break and compromise the entire circuit, and the wires can degrade as they are twisted and stretched. To address this issue, we designed cable routing into the finger links, as seen in the left panel of Fig.~\ref{fig:isyhand_collage}. The first layer of the link (attached directly to a motor) routes the cable from the proximal motor to the underside of the link and from the distal motor to the top. Pre-bending the wires in the primary actuation direction ensures that they will not wear out as quickly. The second layer of the link (shown detached from the motor in Fig.~\ref{fig:isyhand_collage}) covers and protects the wires.\looseness=-1

\subsection{Evaluation}
Similar to Kosanovic and Vaz~\cite{H4ND2024}, we ran multiple experiments  to evaluate the physical capabilities of the \HandName\ as well as its repairability. Camera-based teleoperation was used to evaluate finger speed, progressive loading to evaluate payload capacity, and opposed pressing to evaluate pinching force between the thumb and index and middle fingers. Finally, disassembly and reassembly were performed multiple times for several difficult-to-replace motors. 

\subsubsection*{Finger speed} Finger speed was measured by using MediaPipe~\cite{mediapipe}, a camera-based hand-tracking tool, to teleoperate the \HandName. The hand was fully opened and closed ten consecutive times within a total of 10.32\,s, for an average cycle time of approximately 1\,s.

\subsubsection*{Payload capacity} To measure payload capacity, the \HandName\ was used to repeatedly lift a bucket with progressively increasing weight. The hand was commanded to hold the bucket handle using a power grasp with maximum motor torque, and the experimenter then slowly lifted it. The first lift was performed with 3\,kg, and each successive lift was performed with an additional 0.5\,kg to a maximum of 7\,kg. This process was repeated five times, and the final trial was tested up to 9\,kg. There were zero failures across all lifts. 

\subsubsection*{Pinching force} We placed a scale between the fingertips and ran three pinching tests: one used only thumb and index flexion, the second added the middle finger and palm, and the third added thumb rotation. The mean pinching forces over five trials per test were 3.5\,N, 5.6\,N, and 5.5\,N, respectively.

\subsubsection*{Motor repair time} Five representative joints were disassembled and reassembled by an expert to measure repair time. The average time across all five joints was 5:08 (min.:s), with the outer palm joint taking the longest at 7:26 and the index distal joint taking the shortest at 3:22. 

\section{Simulation Experiments}
To evaluate the in-hand manipulation capability of the \HandName\ compared to the Allegro and LEAP hands as well as the fixed-palm \HandName\ (flat), we perform reinforcement learning (RL) experiments in simulation for the standard in-hand cube reorientation task. Our novel grid-based evaluation systematically evaluates manipulation capabilities across the entire surface of each hand. 

\subsection{Cube Reorientation Experiment Setup}
\label{subsec:cube_reorient}
We compare the \HandName\ (18\,DoF) to a reduced variant with non-articulated palm joints called \HandName\ (flat) (16\,DoF), the Allegro hand (16\,DoF) with BioTac fingertips as set up in~\cite{Makoviychuk21-NIPS-IsaacGym}, and the LEAP hand (16\,DoF) with simulation assets taken from~\cite{shaw2023leaphand} in a cube-reorientation RL task.
A rendering of each hand's simulation asset is shown in Fig.~\ref{fig:all_hands}.

For the cube-reorientation task, we build on the simulation experiment for the Allegro hand in~\cite{Makoviychuk21-NIPS-IsaacGym}. An RL policy is trained with PPO in IsaacGym to manipulate a colored cube to match a target pose (position and orientation); the cube is initialized directly above its target position in an orientation that is the product of consecutive random rotations around the y- and x-axes. Target cube orientations are randomly generated in the same way.
%
We take without modification the original policy's action space, observation space, reward shaping and PPO training hyperparameters from the code base of~\cite{Makoviychuk21-NIPS-IsaacGym} and keep them identical for all hands, although the dimensionality of the action and observation spaces depend on the hand's DoF. We refer to the Allegro hand part of supplementary material of~\cite{Makoviychuk21-NIPS-IsaacGym} for a description of the reward function.
The joint limits (position, velocity and effort) are taken from the respective URDF file of each hand.
While the original joint effort limit for the Allegro hand is set to 0.35 in the URDF, it is increased in~\cite{Makoviychuk21-NIPS-IsaacGym} to 0.5, which we also use for a fairer comparison.
Since there are no actuator-specific properties available in all URDF files and we do not have access to all real hands for measuring the properties, the joint stiffness, damping, friction and armature parameters are set to unified values taken from~\cite{Makoviychuk21-NIPS-IsaacGym}, presented in Table \ref{tab:joint_params}.
All contact-related material properties (friction, restitution, compliance) are the default values of IsaacGym.
To create collision meshes in simulation, the convex mesh decomposition feature of IsaacGym is used with a maximum of four convex meshes per link mesh.
While the translational error of the cube is part of the reward, successes are counted only for matching cube orientations (within a tolerance)~\cite{Makoviychuk21-NIPS-IsaacGym}.

\begin{table}[b] 
\vspace{-6pt}
\caption{Simulation parameters for all joints of all hands.}
\vspace{-5pt}
\label{tab:joint_params}
\begin{center}
\begin{tabular}{lccc}
\toprule
Joint Parameter & \HandName & Allegro Hand & LEAP Hand \\
\midrule
Velocity Limit & 5.50-5.90 & 6.28 & 8.48 \\
Effort Limit & 0.52-0.93 & 0.50 & 0.95 \\
Stiffness & 3.0 & 3.0 & 3.0 \\
Damping & 0.1 & 0.1 & 0.1  \\
Friction & 0.01 & 0.01 & 0.01 \\
Armature & 0.001 & 0.001 & 0.001 \\
\bottomrule
\end{tabular}
\end{center}
\vspace{-5pt}
\end{table}

When running the original task, we found that the number of successful consecutive cube reorientations highly depends on the position of the cube relative to the hand. To test how the performance of each hand depends on the initial/target location of the cube, we trained and evaluated RL policies across a grid of cube positions that extends beyond the entire inside surface of the hands.
We place the origin of the grid's x- and y-coordinates directly above the center of the joint axis that connects the middle finger to the palm (grid-origin cell framed black in Fig.~\ref{fig:grid_results}) at the height of the palm surface.
The grid extends from --10\,cm to 14\,cm horizontally and from --14\,cm to 18\,cm vertically with an equidistant spacing of 2\,cm.
For each grid cell a policy is trained in simulation following the approach in~\cite{Makoviychuk21-NIPS-IsaacGym}, where the initial and target x- and y-position for the cube are set to the respective grid-cell position and the cube's initial and target z-height is fixed to half the cube's diagonal length (5.6\,cm) plus an 0.5\,cm offset above the palm surface. For fairness, at the beginning of each episode, all hands are initialized in a horizontal pose with the fingers laid flat and the thumb pointing upwards.

Like~\cite{Makoviychuk21-NIPS-IsaacGym}, we set the number of training epochs to 5,000 (around 655 million simulation steps) for the grid experiments. This training depth is a reasonable tradeoff between performance and training time, as each hand learns to perform multiple successful cube reorientations, and the average real-world training time across all hands for a single grid cell policy is 3.66 hours on an NVIDIA A40 GPU.
Parallelizing training across multiple GPUs (e.g., on a computing cluster) makes it feasible to train 221 grid cell policies for each hand over multiple seeds.
%
After training all grid-cell policies for 5,000 epochs, we evaluated them on a separate, fixed test sequence of target orientations that were randomly sampled as in training and generated beforehand. This deterministic procedure allows us to compare the performance of each grid-cell policy to any other grid-cell policy of any other hand.
Each grid-cell policy is evaluated on 100 20-s-long episodes for which the number of consecutive cube reorientation successes is counted. We run the full grid training and evaluation with three different seeds for each hand.

Although this experiment rigorously tests early learning for each hand, we did not observe training convergence within 5,000 epochs. Thus, we trained an additional single policy per hand to 300,000 epochs to compare further training progress and performance convergence. The single policy for each hand was trained on the best performing grid-cell (marked green in Fig.~\ref{fig:grid_results}) from the above evaluation using one of the original three seeds. Then, the policy with the highest reward during these 300,000 training epochs was evaluated as above (see Fig. \ref{fig:best_policy_eval}).
Since training for such an extended period takes several days per policy, conducting the full grid experiment for 300,000 epochs would be computationally challenging and expensive, even with access to a large computing cluster.


\subsection{Results and Discussion for Cube Reorientation}

\begin{figure*}[ht]
    \centering
    
    \includegraphics[width=0.24\textwidth,trim={1.5cm 0.5cm 2.0cm 1.5cm},clip]{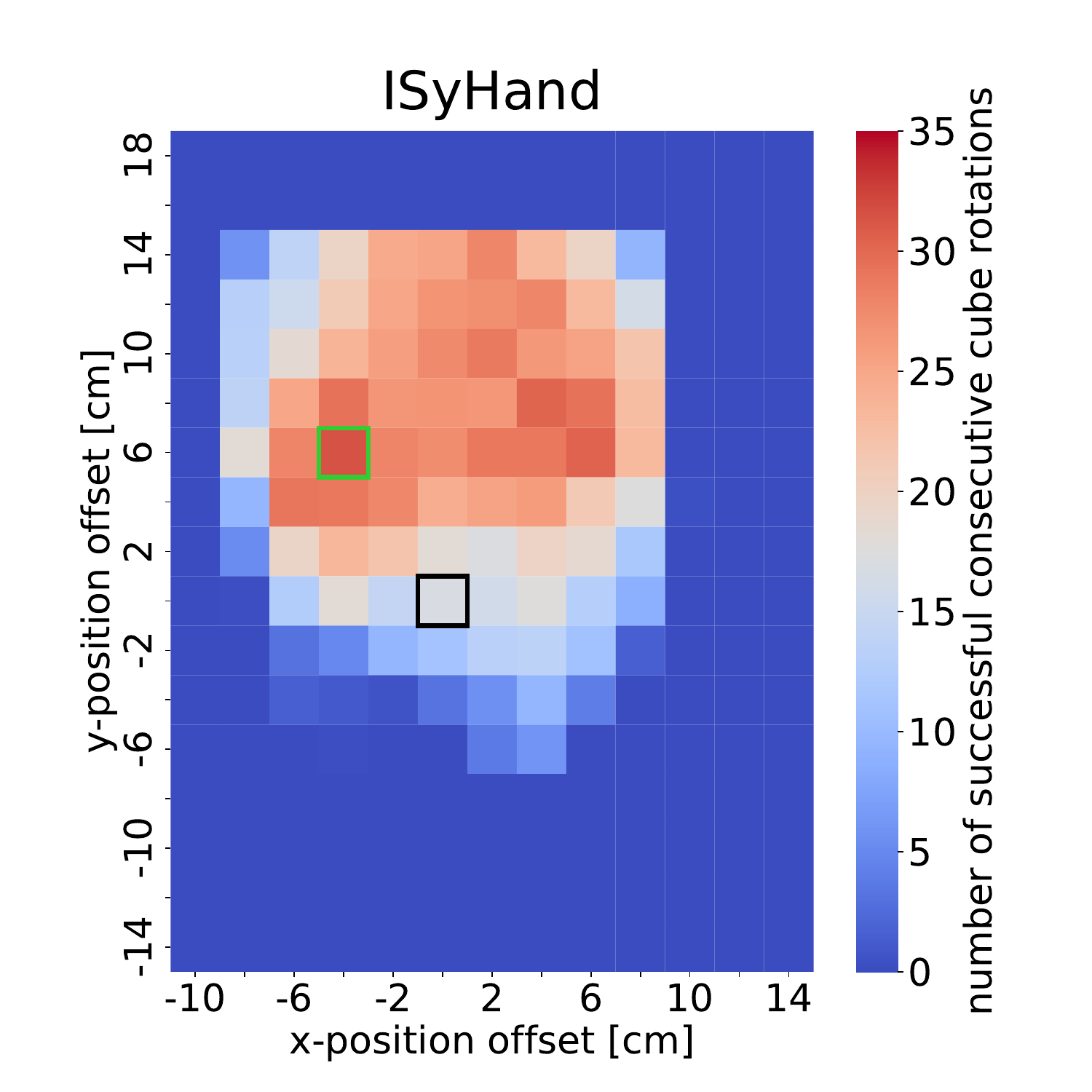}
    \hfill
    \includegraphics[width=0.24\textwidth,trim={1.5cm 0.5cm 2.0cm 1.5cm},clip]{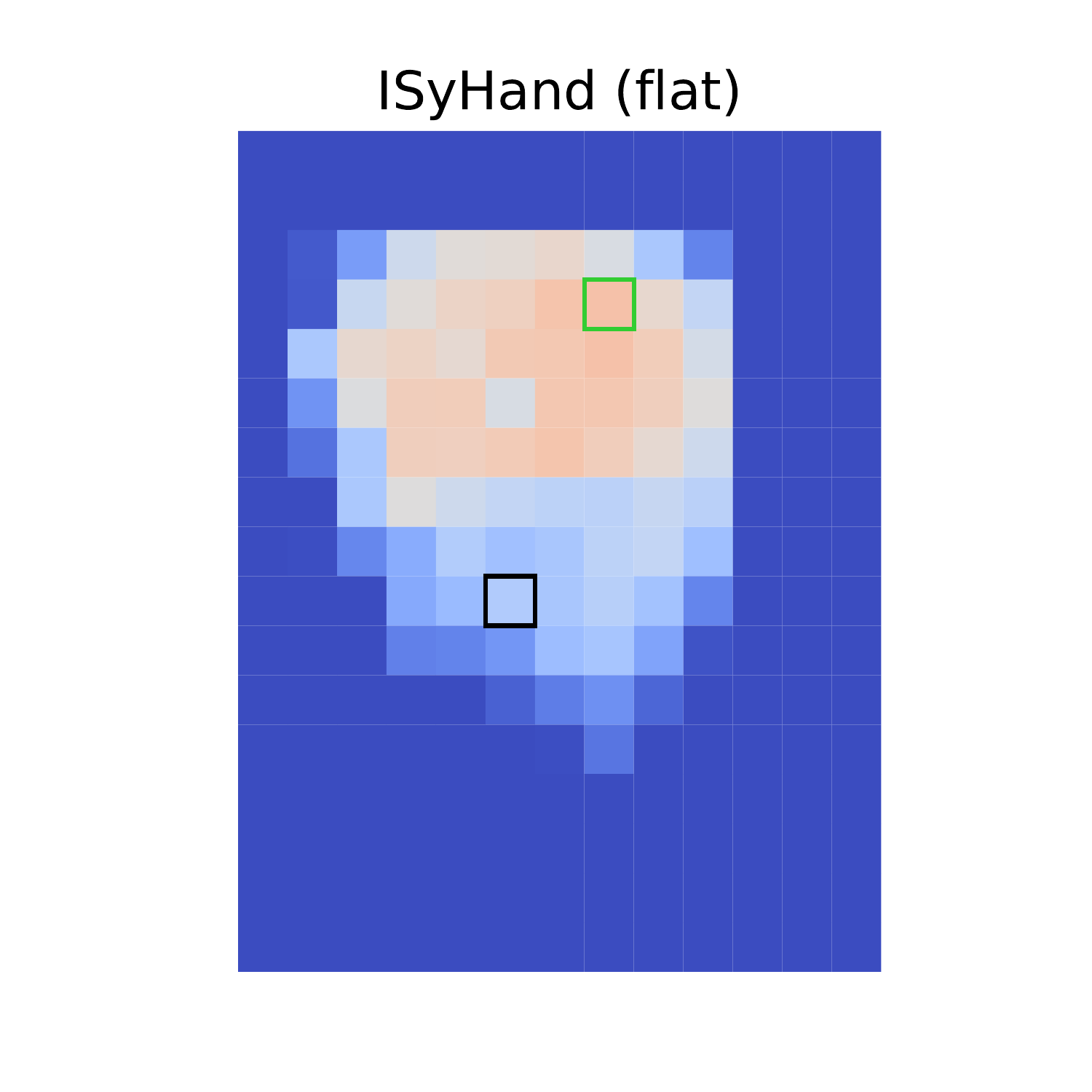}
    \includegraphics[width=0.24\textwidth,trim={1.5cm 0.5cm 2.0cm 1.5cm},clip]{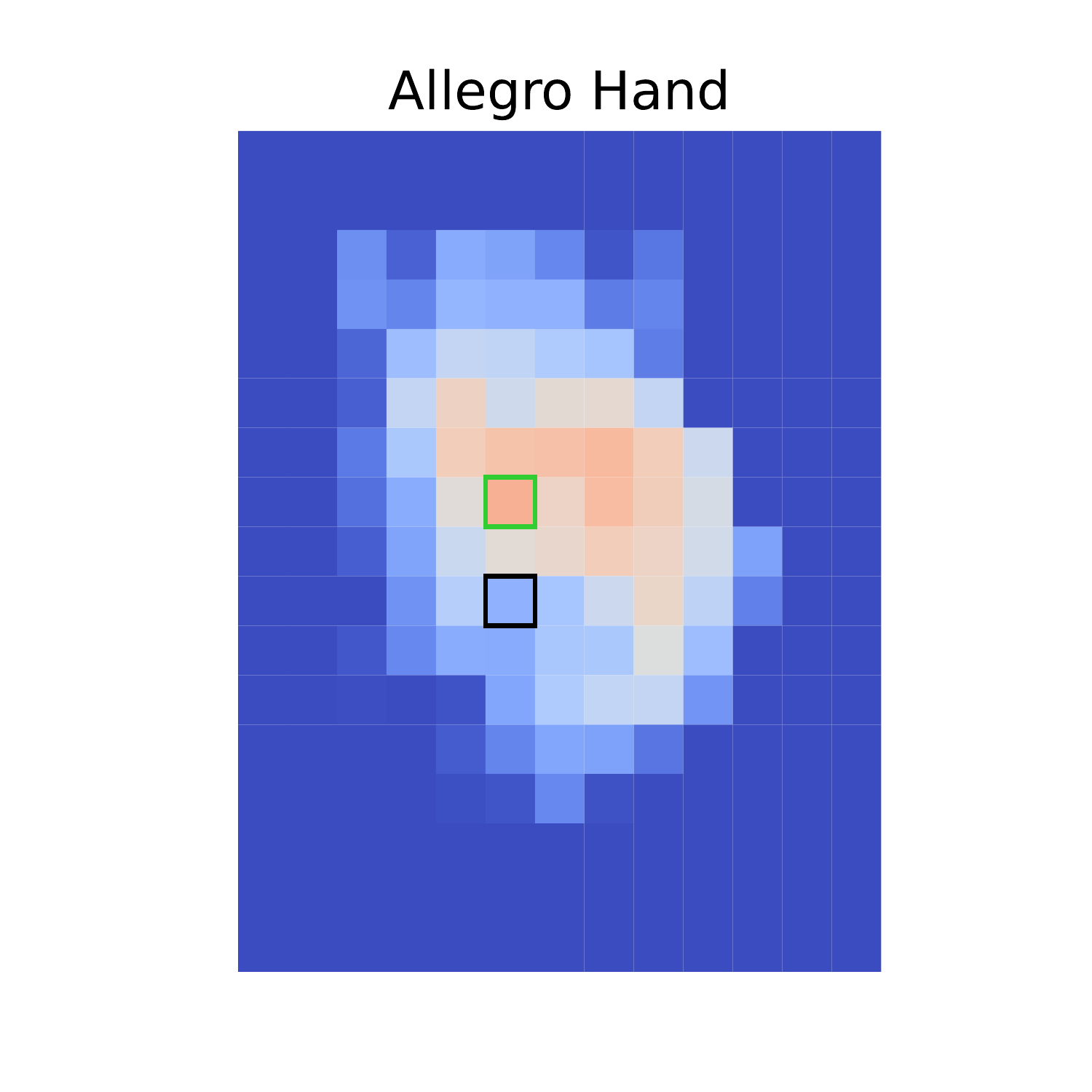}
    \hfill
    \includegraphics[width=0.24\textwidth,trim={1.5cm 0.5cm 2.0cm 1.5cm},clip]{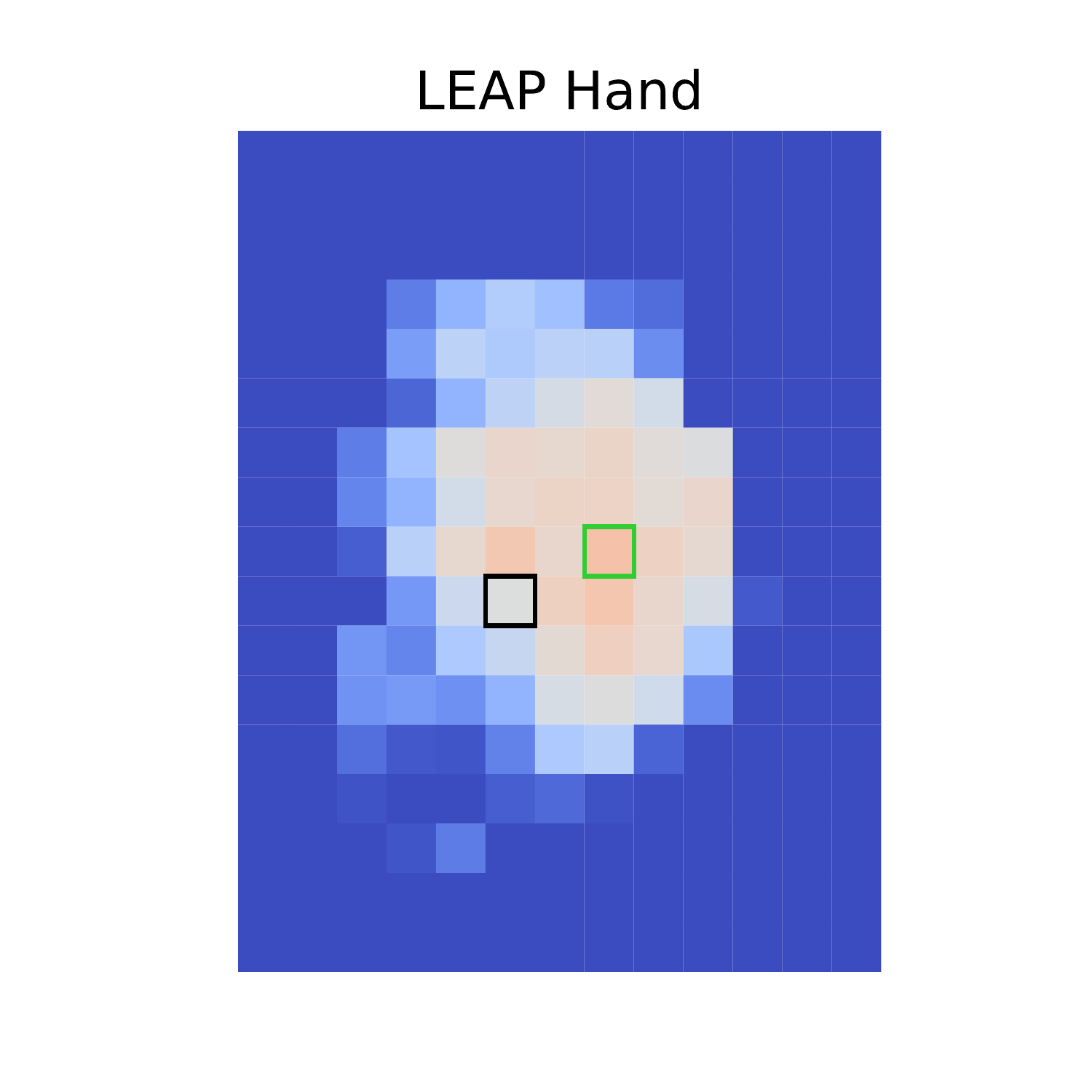}

    
    \caption{Average evaluation performance of each hand on the simulated cube-reorientation task for all grid points, with three policies trained for 5,000 epochs per grid cell. For each hand, the grid cell associated with the most reorientation successes is outlined in green.
    \HandName\ clearly outperforms \HandName\ (flat) and the Allegro and LEAP hand in terms of consecutive rotations and size of manipulability region.
     }
    \label{fig:grid_results}
    \vspace{-10pt}
\end{figure*}

Fig.~\ref{fig:grid_results} displays the evaluation results of the grid-cell policies trained for 5,000 epochs.
The \HandName's policies associated with the best performing grid-cell (marked green in Fig.~\ref{fig:grid_results}) complete 31.46 consecutive cube reorientations on average. The Allegro hand's best grid-cell policies achieve 24.16, followed by the LEAP hand's with 22.27 and the \HandName\ (flat)'s with 22.25 on average. 
In addition to highlighting the best-performing regions of the hands, the grid experiment also reveals how manipulation performance varies across each hand's palmar workspace. Table~\ref{tab:grid_success_thresh} reports the maximum and sum of all grid-cell values and the number of grid-cells whose values surpass certain thresholds. The \HandName\ has the most grid cells that score more than 1, more than 10, and more than 20 successful consecutive cube reorientations. It is also the only hand that achieves more than 30 in any grid cell during this early-learning experiment.

\begin{figure}[tb]
    \centering
    \includegraphics[width=0.45\textwidth, trim={0cm 0cm 0cm 0cm}, clip]{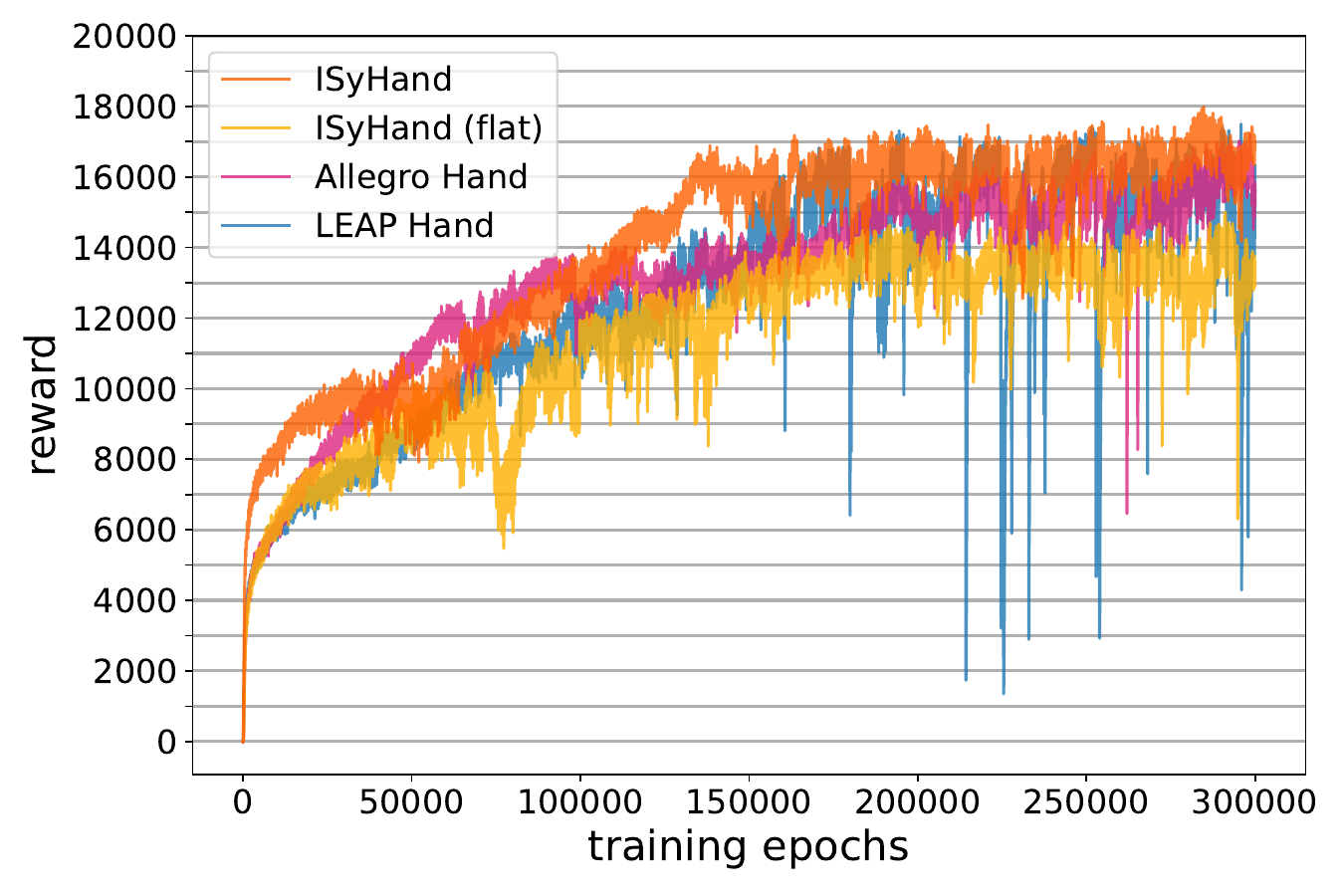}
    \vspace{-7pt}
    \caption{Training curves for each policy trained to 300,000 epochs. The \HandName, Allegro hand, and LEAP hand all achieve similar final performance, outperforming \HandName\ (flat). Training appears to converge around 200,000 epochs.}
    \label{fig:best_policy_training}
    \vspace{-10pt}
\end{figure}

The reward curves for the 300,000-epoch training are shown in Fig.~\ref{fig:best_policy_training}.
While the Allegro and LEAP hands show steady progress throughout training, the \HandName\ achieves the fastest learning within the first 25,000 epochs.
It is also the first hand to reach a reward value close to its convergence plateau ($\sim$16,000 reward at $\sim$150,000 epochs) with only marginal improvement afterward. Toward the end of training, the \HandName, Allegro hand, and LEAP hand reach similar reward ranges of 15,000 to 17,000.
\HandName\ (flat) achieves only a lower reward range between 13,000 and 14,000.

\begin{table}[b] 
\vspace{-10pt}
\caption{The maximum and sum of all grid-cell values and the number of grid cells that hold success values $s$ greater than or equal 1, 10, 20 and 30 for each hand in the simulated in-hand cube-reorientation task. The \HandName\ outperforms the other hands in every category and is the only one that achieves a policy with more than 30 successful consecutive cube rotations on average during the evaluation episodes.}
\vspace*{-2ex}
\label{tab:grid_success_thresh}
\begin{center}
\begin{tabular}{lcccc}
\toprule
 & \HandName & \HandName\ (flat) & Allegro Hand & LEAP Hand \\
\midrule
$\max s$ & $\mathbf{31.46}$ & $22.25$ & $24.16$ & $22.27$ \\
$\Sigma s$ & $\mathbf{1638}$ & $1084$ & $901$ & $923$ \\
\midrule
$s \geq 1$  & $\mathbf{87}$  & $79$  & $78$ & $75$ \\
$s \geq 10$ & $\mathbf{70}$  & $59$  & $42$ & $47$ \\
$s \geq 20$ & $\mathbf{42}$  & $17$  & $10$ & $6$ \\
$s \geq 30$ & $\mathbf{3}$   & $0$   & $0$  & $0$ \\
\bottomrule
\end{tabular}
\end{center}
\vspace{-5pt}
\end{table}

\begin{figure}[tb]
    \centering
    \includegraphics[width=0.45\textwidth, trim={0cm 0cm 0cm 0cm}, clip]{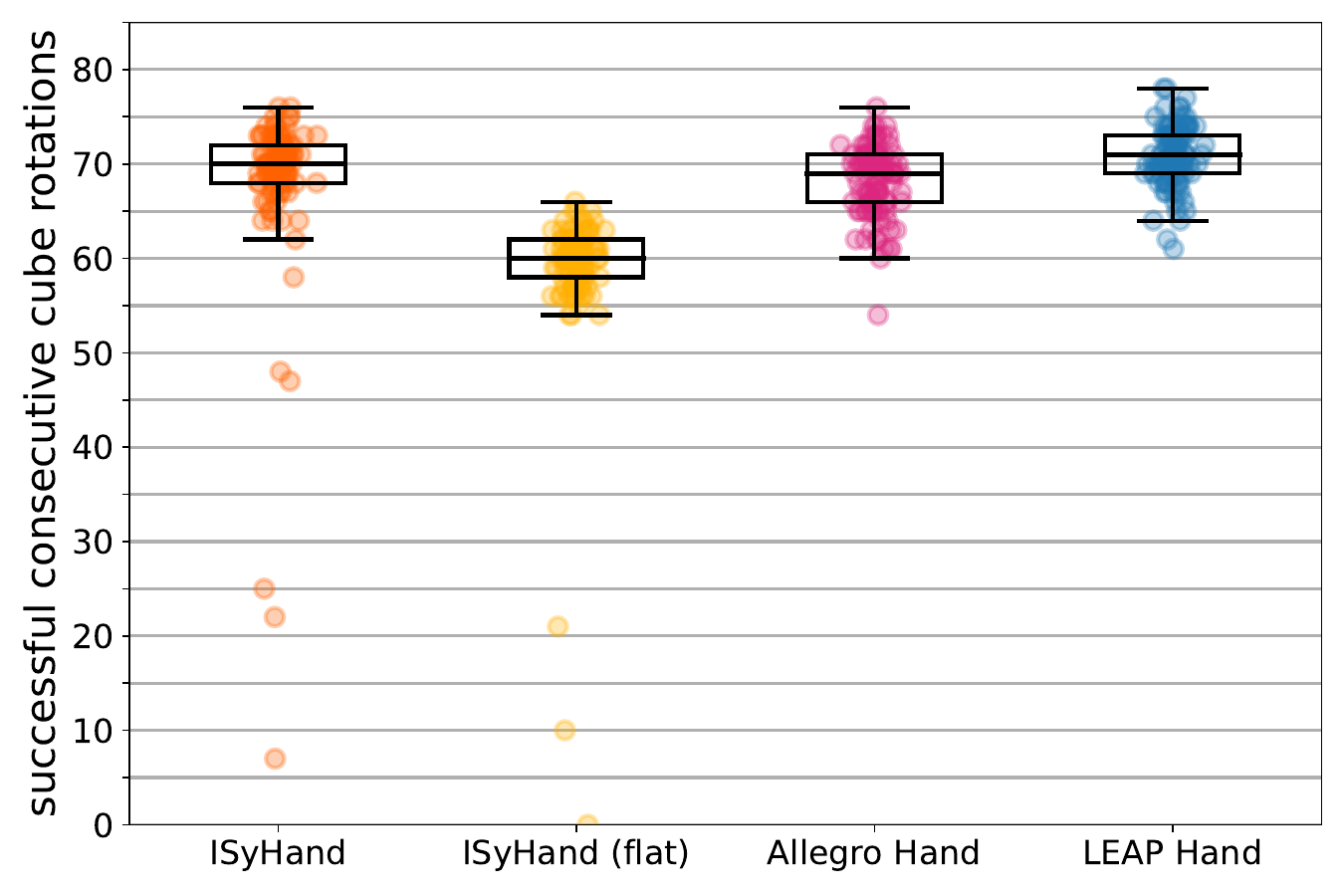}
    \caption{Evaluation performance of the best policy for each hand after 300,000 epochs. The \HandName, Allegro hand, and LEAP hand clearly outperform \HandName\ (flat). All pairwise differences between hands are statistically significant ($p < 0.05$ with a Bonferroni correction).}
    \label{fig:best_policy_eval}
    \vspace{-13pt}
\end{figure}

Fig.~\ref{fig:best_policy_eval} reports the evaluation performance of the policy that achieves the highest reward within the 300,000 training epochs for each hand. The LEAP hand achieves the highest median of successful consecutive cube rotations with 71, closely followed by the \HandName\ with 70, the Allegro hand with 69 and the \HandName\ (flat) with 60.
We applied the paired Wilcoxon signed-rank test on all pairs of hands to assess whether the evaluation samples stem from distributions that are statistically similar. The differences between the distributions are statistically significant with $p<0.05$ after a Bonferroni correction.
Recordings of the best-performing grid-cell policies of each hand for both the early and long-term trainings are shown in the supplementary video.
Due to the approximate modeling of the actuators and the links in simulation, the policies trained in simulation for these experiments are unlikely to run on the real robot hands without performance loss.
Thus, we train a new policy to decrease the sim-to-real gap.




\section{Real-world Experiment}
\begin{figure}[t]
    \centering
    \includegraphics[width=0.9\linewidth]{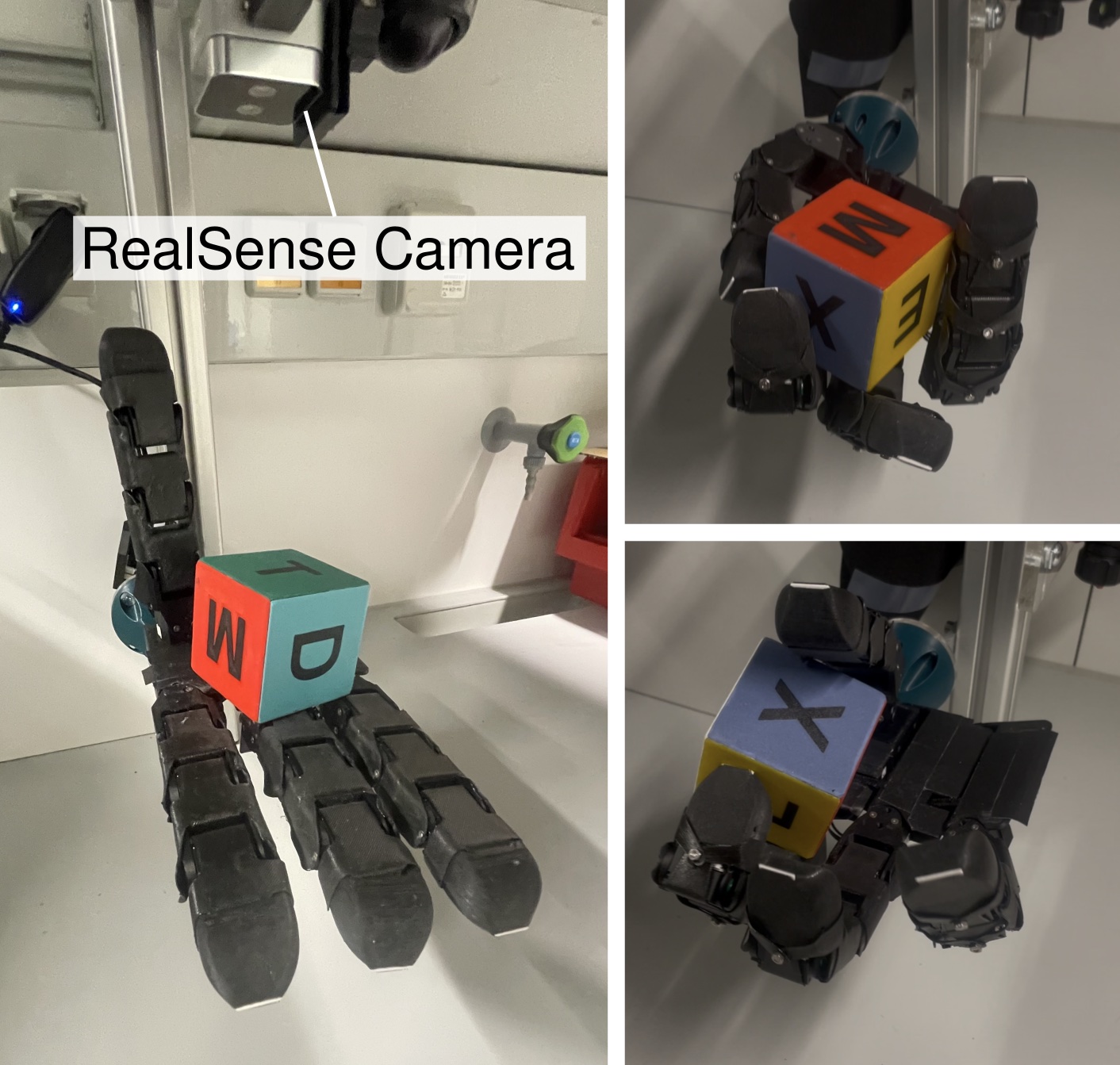}
    \caption{The robot setup with the camera positioned over the hand (left). Two stills taken during a policy rollout (right).}
    \label{fig:isyhand_setup}
    \vspace{-13pt}
\end{figure}

To evaluate the capabilities of the real \HandName, we performed a sim-to-real experiment by training an RL policy in IsaacGym for cube reorientation and deploying it directly on the real robot. We trained the policy for fully random target cube orientations in three axes. We used FoundationPose~\cite{foundationposewen2024} with an Intel RealSense D405 RGB-D camera \cite{intel_realsense} to track cube pose. The setup can be seen in Fig.~\ref{fig:isyhand_setup}. Because both the \HandName\ links and the cube are hard plastic, they have such low contact friction that the cube can slide out of the palm if it is slightly tilted. To solve this mechanical issue, we wrapped the finger links in 3D-printed flexible covers made from Recreus TPU Filaflex 70A and covered the palm in tough tape.\looseness=-1

Because the standard cube-reorientation environment is unlikely to train a policy that would be successful in the real world, we instead trained our policy using DeXtreme \cite{handa2023dextreme}. At a high level, DeXtreme uses curriculum learning to perform automatic domain randomization in simulation to train a policy that is robust to real-world variability and noise. At the beginning of training, the randomization is almost none. As the policy improves and learns the task, more randomization is added. When trained to convergence, the model can learn to deal with a broad range of friction with the object, random noise and latency in both the robot joint readings and commands, noise in the object pose estimate, and more. As such policies are computationally intensive to train, we trained only a single policy on 8,192 agents for 10,000 epochs, much less than the 16,384 agents per GPU across 8\,GPUs on which the original DeXtreme policies were trained~\cite{handa2023dextreme}.\looseness=-1

\subsubsection{Setup}
The robot hand setup consists of three primary components: the object-pose tracker, the robot hand driver, and the policy. Each of these components runs in a separate ROS2 node and communicates over ROS topics. Additionally, we use a 3D-printed version of the cube used in the DeXtreme work. The cube weighs approximately 95\,g and is 6.5\,cm per edge.
As stated above, FoundationPose~\cite{foundationposewen2024} is used to track the position and orientation of the cube from the RealSense RGB and depth images. On an Nvidia GeForce RTX 2080 Ti, the pose estimator runs at approximately 31\,Hz and is fed into the policy. 
The hand driver runs at 100\,Hz, relaying pose commands from the policy to the hand motors and communicating the most recent joint positions to the policy. 
The policy is deployed using ONNX Runtime at 30\,Hz to match the default control rate of the simulation. 

\subsubsection{Experiment}
To evaluate the overall performance of the policy on the \HandName, we roll out the policy 20 times for two minutes per rollout and count the number of consecutive successes. If the cube is dropped, the episode ends. If the hand becomes stuck in an unrecoverable position, the episode continues. Finally, we discard any episode for which the pose-tracking fails, defined as a substantial deviation between the actual and perceived cube poses. 

\begin{table}[b]
    \vspace{-10pt}
    \centering
    \caption{Summary of the cube-reorientation experiments with the real \HandName\ across 30 two-minute-long episodes.}
    \begin{tabular}{ccccc}
    \toprule
        Mean Cube & Max Cube & Drops & Stuck & Pose-Tracking \\
        Reorientations   &  Reorientations &       &       &  Failures \\ 
        \midrule
        6.1 & 16 & 3 & 7 & 10\\
        \bottomrule
    \end{tabular}
    \label{tab:real_results}
\end{table}

\subsubsection{Results and Observations}
The policy was rolled out a total of 30 times with 10 episodes discarded because of a pose tracker failure. An example rollout can be seen in the supplementary video. The results of the episodes are summarized in Table~\ref{tab:real_results}, and two stills from one of the episodes are shown in Fig.~\ref{fig:isyhand_setup}. Overall, the hand performs well considering only a single policy was trained. We observed the palm providing stability and robustness by catching the cube on the outside hand edge and rolling it back onto the palm center. The average number of successful cube reorientations per episode was 6.1, while the fewest and most were 0 and 16. The most common failure case was the pose tracking, which could be remedied by adding additional cameras or training a custom object tracker. A less common but concerning failure case was the cube getting stuck. Typically, this issue occurred when a corner of the cube would catch the edge of the 3D-printed friction wrap. Most of these failures could be remedied with tighter wrapping. We originally also noticed the corners of the cube frequently getting stuck between two fingers or two finger links. Slightly rounding the edges and corners of the cube before testing greatly reduced this issue; increasing the resolution of the contact model in simulation might lead to a better policy. 

It is also important to note that we trained our policy directly on the standard DeXtreme implementation in IsaacGymEnvs. The method could likely be tuned to the \HandName\ to improve overall performance. Additionally, it might be necessary to train with more agents for the policy to converge across the various domain randomizations. However, even with these limitations, this experiment demonstrates that the \HandName\ can perform dexterous manipulation in the real world and rarely drops the cube.

\section{Conclusion and Outlook}
The \HandName\ is a low-cost, robust, highly dexterous robot hand. It is more modular and adaptable than existing on-joint servo-driven hands, as new links can easily be added and the models can be modified. Additionally, typically weak components of existing hands, such as motor cables and joint limits, are carefully integrated into the design and robustified. Importantly, we introduce a novel articulated palm, which was demonstrated through reinforcement learning experiments to improve in-hand manipulation skills. Indeed, our full grid-based cube reorientation experiment 
provides a uniquely rigorous evaluation of the in-hand manipulation capabilities of the \HandName\ and the popular Allegro and LEAP hands, highlighting the benefit of the articulated palm in early training stages. 
This evaluation method could be used for comparisons with concurrent (such as the Tessolo DG-5F hand~\cite{tesollo}) and future hand designs, and similar approaches could be applied to additional tasks. Additionally, the evaluation could be more realistic if hands are trained using DeXtreme-like domain randomization. These types of experiments could shine more light on optimal hand designs, and learning-based approaches could even be used to further optimize hand design, including the location and angle of palm articulations.
Finally, our sim-to-real experiments demonstrate that a policy trained using DeXtreme in simulation can be deployed on the real \HandName.\looseness=-1

\section*{Acknowledgments}
The authors thank the MPI-IS Grassroots program and Cyber Valley for supporting the development of the \HandName. L. Mack and J. Stueckler gratefully acknowledge the HPC resources provided by the Erlangen National High Performance Computing Center (NHR@FAU) of the Friedrich-Alexander-Universität Erlangen-Nürnberg (FAU) under the BayernKI project v119ee for the RL grid experiments in simulation. BayernKI funding is provided by Bavarian state authorities. This work was also partially supported by the Robotics Institute Germany (RIG), which is funded by the German Federal Ministry of Research, Technology and Space (BMFTR) under grant number 16ME1008. 




                              
\bibliographystyle{bib/IEEEtran}
\bibliography{root}

\end{document}